\documentclass[a4paper, 10pt, conference]{ieeeconf}      
\usepackage{FG2026}
\usepackage{url}
\usepackage{algorithm}
\usepackage{algorithmic}
\usepackage{multirow}
\usepackage{amsmath}
\usepackage{tabularx}
\usepackage{adjustbox}
\usepackage[table,xcdraw]{xcolor}
\usepackage{pifont}
\usepackage{subcaption}
\usepackage{wrapfig}      
\usepackage[font=small]{caption} 
\usepackage{lipsum}  
\usepackage[colorlinks=true, urlcolor=magenta]{hyperref}
\FGfinalcopy 

\IEEEoverridecommandlockouts                
\def\FGPaperID{194} 

\title{\LARGE \bf
Open Set Face Forgery Detection via Dual-Level Evidence Collection
}

\author{\parbox{16cm}{\centering
    {\large Zhongyi Cai$^1$, Bryce Gernon$^2$, Wentao Bao$^1$, Yifan Li$^1$, Matthew Wright$^2$, Yu Kong$^1$}\\
    {\normalsize
    $^1$Michigan State University, East Lansing, USA\\ 
    $^2$Rochester Institute of Technology, Rochester, USA}}
    \thanks{
    This material is based upon work supported by NSF grant 2429836 under the NSF Responsible Design, Development, and Deployment of Technologies (ReDDDoT) program, which is jointly sponsored by NSF and the Ford Foundation, The Patrickj. McGovern Foundation, Pivotal Ventures, The Eric and Wendy Schmidt Fund for Strategic Innovation, and Siegel Family Endowment.}
}

\usepackage{fancyhdr}
\begin{document}

\ifFGfinal
\thispagestyle{empty}
\pagestyle{empty}
\else
\author{Anonymous FG2026 submission\\ Paper ID \FGPaperID \\}
\pagestyle{plain}
\fi
\maketitle
\thispagestyle{fancy}

\begin{abstract}
The surge in face forgeries has increasingly undermined confidence in the authenticity of online content. 
As generation algorithms rapidly evolve, new fake categories will constantly emerge, severely challenging existing face forgery detection methods.
Although face forgery detection has recently improved, current techniques remain largely confined to binary Real-vs-Fake classification or the recognition of known fake categories.
Moreover, they fail to identify the emergence of entirely new forgery methods.
In this work, we study the \emph{Open Set Face Forgery Detection (OSFFD)} problem, which requires the detection model to identify novel fake categories. 
To enhance its real-world applicability, we reformulate the OSFFD problem and address it through uncertainty estimation.
Specifically, we propose the Dual-Level Evidential face forgery Detection (DLED) approach, which estimates prediction uncertainty by extracting and integrating category-specific evidence on the spatial and frequency levels.
Comprehensive experiments across diverse settings demonstrate that our proposed DLED approach achieves state-of-the-art performance.
Notably, it surpasses various existing baseline models by a $20\%$ margin on average when identifying forgeries from novel fake categories.
Concurrently, our DLED method yields competitive performance on the standard binary Real-versus-Fake face forgery detection task.
Our codes are publicly available at    \url{https://github.com/zhyczy/DLED}.
\end{abstract}

\section{Introduction}
\label{sec:intro}
Deepfakes, which use deep learning techniques to generate or modify faces and voices, continue to rapidly increase in both sophistication and accessibility.
The diversity of deepfake forgeries \cite{korshunova2017fast, karras2017progressive, shen2017learning, siarohin2019first} causes different visual artifacts to appear in the generated deepfakes, making deepfake detection increasingly difficult.
According to a survey by Mirsky et al.~\cite{mirsky2021creation}, existing face deepfake forgeries can generally be organized into four categories: Face Swapping (FS), Face Reenactment (FR), Entire Face Synthesis (EFS), and Face Editing (FE).
As new generation methods continue to emerge, it is likely that novel categories of facial deepfakes will be developed.

Despite progress in deepfake detection under closed set scenarios \cite{yan2023deepfakebench, F3net, gu2022exploiting, core}, where both training and testing data contain the same known fake forgeries\footnote[1]{In this paper, we define ``fake forgeries" as specific deepfake methodologies, and ``fake categories" as the broader groups to which these methodologies belong; e.g., FSGAN \cite{nirkin2019fsgan} is the fake forgery and Face Swapping is its according fake category.}, these methods have yet to fully address the challenge of generalization to unseen fake forgeries. 
Some studies \cite{wang2020cnn, recce, Nadimpalli_2022_CVPR, uia_vit, CPL} have proposed mechanisms to improve generalization to unseen forgeries. 
However, their overall performance remains suboptimal, and they fail to detect the emergence of novel fake categories.

\begin{figure}[t]
  \centering
  \includegraphics[width=\linewidth]{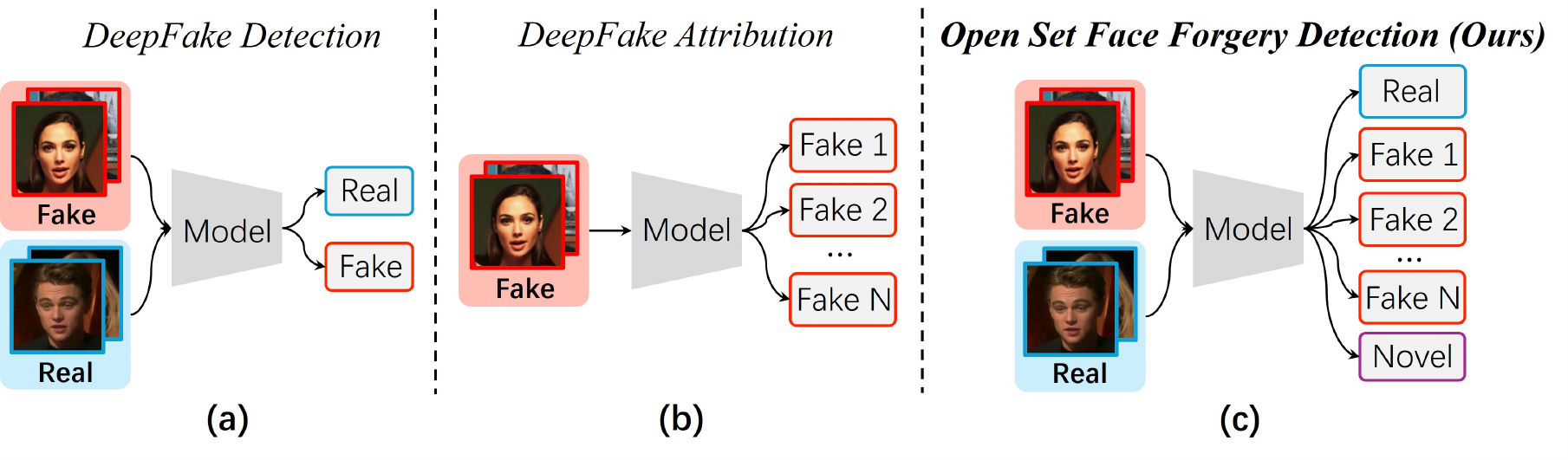}
  \caption{\textbf{Comparison with existing settings}.
  Different from DeepFake Detection (a) and Attribution (b), Open Set Face Forgery Detection (c) aims to identify whether a forgery originates from a novel fake category or not while simultaneously performing multiclass classification among real and known fake categories.}
  \label{fig:setting_comparison}
  \end{figure}

In this paper, we study the Open Set Face Forgery Detection (OSFFD) problem to address this issue. 
OSFFD was proposed in \cite{diniz2024open, zhou2024fine}, but it remains an under-explored problem.
Traditional deepfake detection and attribution tasks either distinguish between real and fake images or assign forgeries to predefined categories. 
In contrast, OSFFD determines whether a given face belongs to a novel fake category, while simultaneously performing multiclass classification among real and known fake categories.
The difference among these settings is shown in Fig. \ref{fig:setting_comparison}.
The aforementioned studies approached the OSFFD problem by training models on labeled data for seen classes (real and known fake categories) and unlabeled data for novel fake categories.
This setup has practical limitations as data from a novel fake category would not be integrated into datasets immediately after its proliferation.
In this paper, we reformulate the OSFFD problem by restricting model training to only real and known fake categories, which enhances the real-world applicability of OSFFD. 

To address the OSFFD problem, we formulate it as an uncertainty estimation issue that assesses the confidence of model predictions based on the evidence collected from the data.
During training, the model is exposed to known fake categories and learns to assign low uncertainty to these samples. 
At test time, samples from unknown categories are expected to yield high uncertainty scores, facilitating their detection. 

Furthermore, we propose a novel \emph{Dual-Level Evidential face forgery Detection} approach, DLED, that simultaneously identifies emerging, unknown fake categories and performs multiclass classification among real and known fake categories.
To enable novel category recognition, DLED leverages Evidential Deep Learning (EDL) \cite{sensoy2018evidential, sensoy2020uncertainty, shi2020multifaceted} for classification and uncertainty estimation.
However, unlike conventional open set classification, OSFFD operates on structured facial imagery whose spatial semantic patterns alone are insufficiently discriminative~\cite{wang2020cnn}. 
Accordingly, DLED augments these cues with complementary low-level frequency artifacts, yielding a more effective application of EDL.
Since both sources are informative, detection decisions should reflect their joint support. 
To this end, we introduce an uncertainty-guided evidence fusion mechanism grounded in Dempster’s combination rule~\cite{sentz2002combination}, enabling DLED to integrate evidence on both the spatial and frequency levels into a unified, comprehensive uncertainty estimate.
Furthermore, 
we propose an improved uncertainty estimation approach to enhance the model’s capability to detect novel fakes, as the original EDL formulation
can be affected by evidence from irrelevant classes, resulting in suboptimal uncertainty quantification.

Compared with existing face forgery detection methods, a key advantage of our DLED model lies in its ability to promptly detect newly emerging fake categories and avoid misclassification, without relying on any prior knowledge of these categories. 
While existing deepfake detection algorithms can be adapted to be feasible for the OSFFD problem, e.g., one-class detectors~\cite{sbi, oc-dect, larue2023seeable} can combine with a separate multiclass classifier, they often struggle to balance between accurate novel category detection and effective multiclass classification.
In addition, our methodology is grounded in principled reasoning, offering clear interpretability for the OSFFD results.

In summary, our contribution is three-fold:
\begin{itemize}
\item 
We reformulate the Open Set Face Forgery Detection (OSFFD) problem, eliminating the reliance on unlabeled data from novel fake categories during model training, making it more applicable in real life.

\item 
We propose leveraging EDL to treat the OSFFD problem as an uncertainty estimation problem, enabling the model to determine whether a face image originates from a novel fake category.

\item 
We propose the DLED approach, which aggregates and fuses evidence on both the spatial and frequency levels to estimate prediction uncertainty. Extensive empirical results validate its effectiveness and demonstrate its superiority over various baseline models.
\end{itemize}

\section{Related Work}
\noindent{\bf Deepfake Detection.}
A wide range of deepfake detection approaches have been studied in the literature \cite{huang2022towards}. 
Most existing methods \cite{sia, core, uia_vit, recce} leverage spatial patterns to detect manipulation artifacts, while others \cite{srm, gu2022exploiting, zhang2019detecting} exploit discrepancies in the frequency domain to reveal forgery traces.
Some studies \cite{npr, dire, trufor} also integrated features from complementary modalities, such as noise patterns, to further distinguish fake faces.
One-class anomaly detection methods \cite{oc-dect, sbi, larue2023seeable} treat real faces as the positive class and all other data as anomalous outliers, training the model exclusively on the positive class to distinguish between real and fake faces.
Recent works \cite{UniFD, prompt_tuning} find that the pretrained CLIP \cite{clip} model performs well on unseen forgeries.
Based on this finding, several recent works \cite{FAA, CFPL-FAS, D3} designed diverse adaptations for CLIP to enhance its detection capabilities.
However, these approaches are limited by their exclusive focus on Real-vs-Fake classification, which overlooks the differences among various fake categories.

\noindent{\bf Deepfake Attribution.}
The deepfake attribution task aims to identify the source of fake faces so that models can provide persuasive explanations for the results of deepfake detection.
However, most of these methods \cite{wu2024traceevader, huang2023can, yang2022deepfake, zhong2023copyright} are limited to the closed set scenario.
Few methods have utilized the ``open world" setting to track unseen forgeries.
The open-world GAN \cite{Girish2021TowardsDA} method is designed to detect images generated by previously unseen GANs, but its framework does not extend to other manipulations such as face editing.
Another work, CPL \cite{CPL}, introduced a benchmark which encompasses a broader array of unseen forgeries derived from multiple known categories. 
However, this setting relies on access to unlabeled data from such forgeries during training and does not determine whether a given forgery originates from a novel category, thereby limiting its practical applicability.
Although recent works~\cite{wang2024bosc, wang2023open} introduced open set classification for forgeries, their settings do not differentiate between unseen forgeries originating from known categories and those from entirely novel categories, nor can they determine whether a face is real or fake.

\noindent{\bf Open Set Recognition.}
Open Set Recognition is a well-defined task that recognizes known classes and differentiates the unknown. 
The pioneering work \cite{scheirer2012toward} formalized the definition and introduced a ``one-vs-set" machine based on a binary SVM.
Prototype learning and metric learning methods \cite{chen2021adversarial, yang2020convolutional, zhang2021prototypical} have been applied to identify the unknown by keeping unknown samples at large distances from prototypes of known class data.
Recently, uncertainty estimation methods \cite{wang2021energy, bao2021evidential, fan2024evidential, flexible_vis} using Evidential Deep Learning (EDL) have shown promising results on open set recognition problems.
EDL \cite{sensoy2018evidential, sensoy2020uncertainty, shi2020multifaceted} works well to quantify model confidence and prediction uncertainty, exhibiting high efficacy in handling unseen data types, and it has been further broadened to encompass multi-view classification \cite{han2020trusted, huang2024crest}. 
To the best of our knowledge, this paper is the first to integrate EDL into the OSFFD problem.
\section{Open Set Face Forgery Detection}
\noindent{\bf Definition.}
\begin{figure}[t]  
  \centering
  \includegraphics[width=0.9\linewidth]{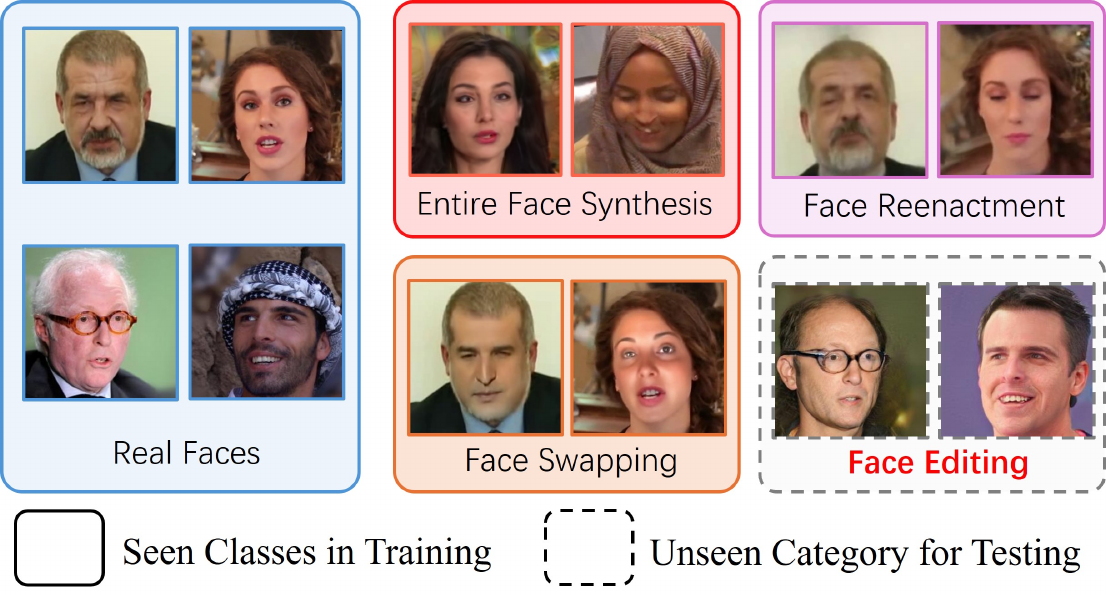}
  \caption{{\bfseries Illustration for Fake Categories in OSFFD.}
Real faces and fake faces from the seen categories are used to train the model. 
Subsequently, the model is evaluated on test data that includes both seen classes and previously unseen categories.
In the figure, the labels EFS, FR, and FS denote seen categories, whereas FE represents an unseen category.}
  \label{fig:category}         
\end{figure}
As depicted in Fig. \ref{fig:category}, Open Set Face Forgery Detection (OSFFD) addresses a practical problem: leveraging knowledge from seen classes (i.e., real faces and faces from known fake categories) to classify a given face as either belonging to a seen class or to the newly emerging, unseen fake category.
In the training phase, the model is exposed exclusively to images from seen classes, while images from novel fake categories are reserved for testing purposes. 

\noindent{\bf Motivation.}
OSFFD requires a model to simultaneously discover novel fake categories and perform multiclass classification.
Among these two objectives, novel fake category discovery is the core challenge. 
However, most existing detectors~\cite{UniFD, lsda} emphasize out-of-distribution (OOD) generalization, which target binary real-vs-fake discrimination on unseen testing samples.
As a result, they neither support multiclass classification nor distinguish novel fake categories, rendering them unsuitable for OSFFD.
One alternative is a two-stage pipeline that first partitions samples into seen versus unseen class via OOD detection~\cite{oc-dect, sbi} and then applies a face-forgery classifier to seen classes;
but this decoupled design optimizes different training objectives across stages and offers limited theoretical interpretability.
Additionally, existing open set recognition (OSR) methods~\cite{zhang2023decoupling, lang2024coarse} could hardly perform well when directly applied to OSFFD as the data in OSFFD consists of highly structured facial imagery that requires additional mechanisms to extract discriminative representations.
Therefore, novel algorithms need to be developed to address the OSFFD problem.

\noindent{\bf Formulation.} 
Given a labeled training set $D_S = \{(x_i, y_i)\}_{i=1}^{M}$ consisting of $M$ labeled samples from $K$ seen classes comprising the Real class and $N$ known fake categories (i.e., $K = N + 1$, $y_i \in \{1, \dots, K\}$) and a test set $D_T$ containing samples from the face class set $\{R, F_1, \dots, F_N, F_{N+1}, \dots, F_{N+U}\}$, where $U$ is the number of unknown fake categories, we denote the
embedding space of class $k\in[1,K]$ as $P_k$, and its corresponding open space as $O_k$.
The open space is further divided into two subspaces: the positive open space from other known classes $O_k^{\rm pos}$ and the negative open space $O_k^{\rm neg}$ that represents the remaining infinite unknown region. 

For a single class $k$, the samples $D_S^k\in P_k$, $D_S^{\neq k}\in O_k^{\rm pos}$, and $D_V\in O_k^{\rm neg}$ are positive training data, negative training data, and potential unknown data respectively.
Then, we could use a simple binary classification model $\Psi_k(x)\rightarrow \{0,1\}$ to detect unseen classes \cite{chen2021adversarial} and optimize the model by minimizing the expected risk $R^k$:
\begin{equation}
\label{eq:binary_risk}
\mathop{\arg\min}_{\Psi_k} R^k= R_{c}(\Psi_k, P_k\cup O_k^{\rm pos})+\alpha\cdot R_o(\Psi_k, O_k^{\rm neg}),
\end{equation}
where $\alpha$ is a positive constant, $R_c$ is the empirical classification risk on the known data, and $R_o$ is the open space risk \cite{scheirer2012toward}. 
$R_o$ measures the likelihood of labeling unknown samples as either known or unknown classes, expressed as a nonzero integral function over the space $O_k^{\rm neg}:$
\begin{equation}
    R_o(\Psi_k, O_k^{\rm neg})=\frac{\int_{O_k^{\rm neg}}\Psi_k(x)dx}{\int_{P_k\cup O_k}\Psi_k(x)dx}.
\end{equation}
The more frequently the negative open space $O_k^{\rm neg}$ is labeled as positive, the higher the associated open space risk.

We extend the single-class detection to the multiclass OSFFD setting by integrating multiple binary classification models $\Psi_k$ using a one-vs-rest strategy.
With (\ref{eq:binary_risk}), the overall expected risk is computed as the sum over all seen classes: $\sum_{k=1}^{K} R^k$.
This is equivalent to training a multiclass classification model $\mathcal{F} = \odot(\Psi_1, \dots, \Psi_K)$ for $K$-class classification, where $\odot(\cdot)$ denotes the integration operation.
The overall training optimization objective is formulated as:
\begin{equation}
\label{eq:opt_obj}
    \mathop{\arg\min} \{R_{c}(\mathcal{F}, D_S)+\alpha\cdot\sum_{k=1}^{K} R_o(\mathcal{F}, D_V)\},
\end{equation}
which demands the model to minimize the combination of the classification risk on seen classes and the open space risk on unseen classes.
Therefore, our goal is to train a multiclass classification model $\mathcal{F}(\cdot)$, parameterized by $\theta$, on $K$ seen classes to accurately classify faces as either real or belonging to one of the known fake categories, while simultaneously detecting novel fake categories as a distinct $(K+1)^\text{th}$ class.
We further formulate OSFFD as an uncertainty estimation problem: the model $\mathcal{F} : \mathcal{X} \rightarrow (\tilde{y}, \tilde{u})$ outputs a predicted category label $\tilde{y}\in\{1,\dots,K\}$ and its associated predictive uncertainty $\tilde{u}$.
If the predictive uncertainty exceeds the class-specific threshold $\tau_{\tilde{y}}$, i.e., $\tilde{u} > \tau_{\tilde{y}}$, the predicted label is deemed unreliable and the instance is assigned to the novel fake category.

\section{Methodology}
To solve the formulated uncertainty estimation problem, we utilize established techniques such as MaxLogit and Evidential Deep Learning.

\noindent{\bf Plug-in OSR Techniques.} Maximum Softmax Probability \cite{hendrycks2019scaling} and MaxLogit \cite{wang2022vim} detectors are two widely used plug-in OSR techniques, which utilize the maximum Softmax probabilities and the maximum logits as the model prediction confidence with no extra computational cost.

\noindent{\bf Evidential Deep Learning.}
Evidential Deep Learning (EDL) is an effective technique that performs multiclass classification and uncertainty modeling by introducing the framework of Dempster-Shafer Theory \cite{sentz2002combination} and subjective logic \cite{josang2016subjective}.
For a $K$-class classification problem, given a sample $x$ and a model $\mathcal{F}$ parameterized by $\theta$, the predicted evidence is given by $e = h(\mathcal{F}(x; \theta)) \in R^{K}$, where $h$ is an evidence function.
With total strength $S=\sum_{k=1}^{K} \alpha_k$, where $\alpha_k = e_k + 1$, the predicted probability for class $k$ is $p_k=\alpha_k/S$ and the prediction uncertainty $u$ is calculated as $u=K/S$. 
EDL has been useful to detect data from unknown classes in prior literature~\cite{bao2021evidential,Zhao_2023_CVPR,yu2024anedl,wang2024towards,peng2025advancing}.
These studies motivates us to develop an EDL-based algorithm to detect novel deepfake categories.
Compared with plug-in OSR techniques, EDL provides a more principled uncertainty estimation.

\begin{figure*}[t]
\centering
  \includegraphics[width=0.86\linewidth]{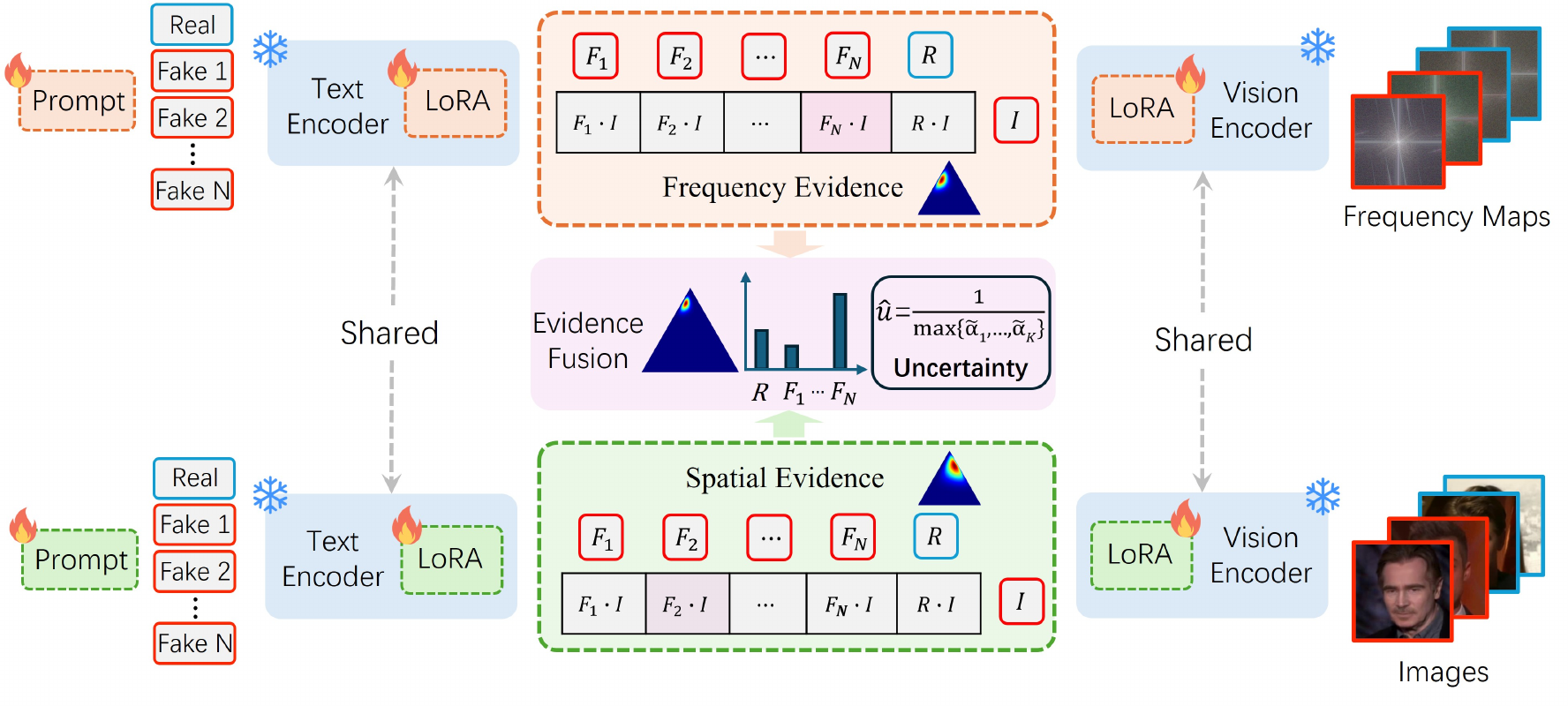}
  \caption{
  {\bfseries Overview of DLED.} 
  DLED collects and fuses evidence from both the spatial and frequency domains to estimate prediction uncertainty. 
  Our improved uncertainty estimation $\hat{u}$ is applied to achieve better detection performance.
  $F_N$ represents the $N$-th fake category and $K$ is the total known class number.
  If the uncertainty for the given sample is larger than the computed threshold, its label will be reassigned to the novel fake category.
  In the evidence illustration, we present a demonstration of a three-class classification scenario ($K=3$).
  }
  \label{fig:main_fig}
\end{figure*}
\noindent{\bf Challenges in applying EDL.}
In our approach, we employ EDL to collect evidence for face forgery detection.
However, leveraging EDL to address the OSFFD problem meets the following challenges:

1) How to collect sufficient evidence in the OSFFD problem?
Unlike conventional open set image classification, face forgery detection involves highly structured facial imagery. 
As a result, off-the-shelf EDL does not directly carry over to OSFFD with satisfactory performance.
To bridge this gap, we extract evidential cues at two complementary levels: high-level semantic signals in the spatial domain and low-level artifacts in the frequency domain. 

2) How can we achieve a comprehensive integration of collected complementary evidential cues?
As both sources carry informative evidence, detection decisions should account for their joint contribution.
The key challenge, therefore, is integrating the two independent uncertainty estimates into one well-calibrated and comprehensive metric.
We address this issue by proposing a novel uncertainty-guided evidence fusion mechanism.
\section{Dual Level Evidence Collection}
\noindent{\bfseries Overview.} 
To address the OSFFD problem, we propose the Dual-Level Evidential face forgery Detection (DLED) approach, which is exhibited in Fig. \ref{fig:main_fig}.
DLED exploits EDL through a dual-level evidential architecture that captures category characteristics of facial imagery across the spatial and frequency domains, yielding sufficiently discriminative evidence. 
It addresses the evidence aggregation challenge with an uncertainty-guided fusion mechanism and further incorporates an uncertainty-improvement procedure to enhance the reliability of the resulting estimates. 
Together, these components enable DLED to detect novel fake categories by quantifying classification uncertainty across complementary levels and determining whether an existing prediction should be reassigned to the novel category. 

\noindent{\bf Spatial and Frequency Evidence.}
\label{sec:csu}
Our DLED model addresses the evidence collection problem by extracting cues at two complementary levels: high-level spatial semantic signals and low-level frequency artifacts.
Face forgeries generally fall into several common categories (FS, FR, EFS, and FE) based on their characteristics in the context of human visuals \cite{mirsky2021creation}.
We refer to these characteristics as deepfake category semantics, which are neglected by most existing works.
Exploiting these semantics, the model can discern subtle differences among fake categories.
To leverage both contextual and visual deepfake semantics, we employ the CLIP \cite{clip} architecture, a vision-language model designed to align image and text representations in a shared semantic space.
Given an input image and the class textual descriptions, we then calculate the logit mass $m_i$ for class $i$. 
In contrast to standard open set classification, relying solely on visual semantics fails to capture the structure of forgery images. 
We thus leverage low-level artifacts in the frequency domain as a complementary source of evidence.
Specifically, for each input image, we obtain its frequency map by applying the Fast Fourier Transform and shifting the resulting spectrum to center the low-frequency components, thereby making them more prominent.
To extract evidence from these complementary domains, we employ two parallel CLIP pipelines, each with a dedicated image encoder and text encoder.
Since CLIP is not explicitly trained to capture forgery image patterns, particularly in the frequency domain, we adapt it by fine-tuning the encoders along with the text prompts while freezing all other pretrained parameters.
For text prompts, we employ Context Optimization~\cite{coop}, which augments class tokens with learnable prompt vectors to yield stronger context embeddings.
For the image and text encoders, we integrate LoRA~\cite{lora} layers into them, which enhance the model's understanding of deepfakes while not adding any additional parameters during testing.
Although we have two parallel branches for spatial and frequency level representations, we reduce memory consumption by sharing their pretrained parameters.

\noindent{\bf Evidential Uncertainty Estimation.}
Our DLED model detects novel fake categories through evidential uncertainty estimation using Evidential Deep Learning (EDL)~\cite{sensoy2018evidential} in an end-to-end manner grounded in solid theoretical principles.
EDL employs deep neural networks to output the parameters of a Dirichlet distribution over class probabilities, which is then used for both class prediction and uncertainty estimation. 
This process can be regarded as an evidence collection process.
By leveraging EDL, our method quantifies the uncertainty associated with each prediction to assess its reliability. 
If the uncertainty is high, the model will reclassify the input as belonging to the novel class, thereby enabling the identification of faces from previously unseen fake categories.

Specifically, for each of the spatial and frequency branches with classification logits mass $m$, our approach calculates the corresponding evidence $e=h(m)$ using an evidence function $h(\cdot)$ that guarantees $e$ to be non-negative. 
During the training phase, to facilitate evidence collection, we independently apply the following EDL loss to each branch:
\begin{equation}
    \mathcal{L}_{EDL}(e\,,\, y)=\sum_{k=1}^{K}y_{k}(log S-log(e_{k}+1)),
\end{equation}
where $S=\sum_k \alpha_k$ and $\alpha_k=e_k+1$, denoting the total strength of the Dirichlet distribution governed by $\{\alpha_{1,\ldots,K}\}$, and $y$ is the one-hot $K$-class label. 
We also apply AvU regularization \cite{bao2021evidential, hammam2022predictive} to each branch for uncertainty calibration.
The EDL loss and AvU regularization minimize $R_c$ and $R_o$ in (\ref{eq:opt_obj}) respectively.

\begin{table*}[htb]
\footnotesize
\setlength{\tabcolsep}{3.8mm}
\centering
\caption{
Comparisons of model performance with diverse baseline methods implemented by ourselves for the OSFFD problem. We use different data configurations for the seen and unseen fake categories. For ``FS", ``FR", and ``EFS", we let each fake category be the unseen category and let the remaining two be seen categories. For ``FE \& SM", we take FS, FR, and EFS as seen categories and let FE and SM be the unseen categories. The best results are highlighted in \textbf{bold}.
}
\label{tab:main_OSFFD}
\begin{adjustbox}{width=\textwidth}
\begin{tabular}{cl|cc|cc|cc|cc|cc}
\hline
\multicolumn{2}{c|}{\multirow{2}{*}{Methods}}                                & \multicolumn{2}{c|}{FS}           & \multicolumn{2}{c|}{FR}           & \multicolumn{2}{c|}{EFS}          & \multicolumn{2}{c|}{FE \& SM}     & \multicolumn{2}{c}{Avg}           \\ \cline{3-12} 
\multicolumn{2}{c|}{}                                                        & Acc            & DR               & Acc            & DR               & Acc            & DR               & Acc            & DR               & Acc            & DR               \\ \hline
\multicolumn{1}{c|}{\multirow{2}{*}{Two-stage}}  & OC-FakeDect \cite{oc-dect}              & 58.16          & 14.68            & 60.69          & 11.43            & 56.14          & 9.01             & 56.74          & 11.67            & 57.93          & 11.70            \\
\multicolumn{1}{c|}{}                            & SBI \cite{sbi}                      & 65.15          & 1.07             & 64.19          & 3.00             & 61.24          & 0.91             & 62.27          & 0.66             & 63.21          & 1.41             \\ \hline
\multicolumn{1}{c|}{\multirow{5}{*}{CNN-based + OSR}}  & Xception \cite{ff++}                 & 64.60          & 23.90            & 53.51          & 29.06            & 57.62          & 22.70            & 55.28          & 29.04            & 57.75          & 26.17            \\
\multicolumn{1}{c|}{}                            & SPSL \cite{spsl}                     & 65.07          & 16.71            & 54.10          & 18.93            & 59.67          & 18.12            & 60.02          & 25.98            & 59.71          & 19.93            \\
\multicolumn{1}{c|}{}                            & SIA \cite{sia}                      & 62.09          & 13.59            & 54.62          & 13.36            & 56.85          & 10.99            & 56.29          & 22.53            & 57.46          & 15.12            \\
\multicolumn{1}{c|}{}                            & UCF \cite{ucf}                      & 65.08          & 0.30             & 50.98          & 0.20             & 52.95          & 1.28             & 52.69          & 1.80             & 55.42          & 0.89             \\
\multicolumn{1}{c|}{}                            & NPR \cite{npr}                      & \textbf{75.37} & 17.37            & 64.63          & 6.75             & 70.43          & 4.36             & 71.45          & 29.20            & 70.47          & 14.42            \\ \hline
\multicolumn{1}{c|}{\multirow{5}{*}{CLIP-based + OSR}} & CLIP Closed Set Finetuning & 67.24          & \textbackslash{} & 65.19          & \textbackslash{} & 64.53          & \textbackslash{} & 66.24          & \textbackslash{} & 65.80          & \textbackslash{} \\
\multicolumn{1}{c|}{}                            & CLIP Zero-Shot \cite{clip}            & 52.30          & 0.81             & 50.36          & 0.26             & 46.01          & 0.38             & 47.62        & 0.25             & 49.07          & 0.43             \\
\multicolumn{1}{c|}{}                            & UnivFD \cite{UniFD}                   & 68.81          & 3.88             & 64.00          & 2.48             & 63.21          & 0.73             & 66.34          & 8.22             & 65.59          & 3.83             \\
\multicolumn{1}{c|}{}                            & CLIPing \cite{prompt_tuning}                  & 66.44          & 14.38            & 62.41          & 6.09             & 61.29          & 4.92             & 66.26          & 19.27            & 64.10          & 11.16            \\
\multicolumn{1}{c|}{}                            & $D^3$ \cite{D3}                  & 70.46          & 8.14            & 64.71          & 8.90             &  61.65         &    1.17          &      66.33     &    8.26         &  65.79         &   6.62          \\
\hline
\multicolumn{1}{c}{}                            & \textbf{Ours}             & 71.37          & \textbf{33.61}   & \textbf{66.83} & \textbf{34.92}   & \textbf{75.52} & \textbf{34.71}   & \textbf{74.48} & \textbf{82.18}   & \textbf{72.05} & \textbf{46.35}   \\ \hline
\end{tabular}
\end{adjustbox}
\end{table*}

\noindent{\bf Test-time Evidence Fusion.} 
To address the integration problem, we design an uncertainty-guided test-time evidence fusion mechanism.
During model inference, according to EDL \cite{sensoy2018evidential}, the probabilities of different classes (belief masses) and the overall uncertainty mass can be calculated as $b_k = e_k/S$ and $u=K/S$.
The $K$ belief mass values and the uncertainty $u$ are all non-negative and follow the sum-to-one rule: $\sum_{k=1}^K b_k+u=1$.  With this approach, we can obtain the belief and uncertainty for each branch.

Our DLED model collects two independent sets of probability mass $M^s=\{\{b_k^s\}_{k=1}^K, u^s\}$ and $M^f=\{\{b_k^f\}_{k=1}^K, u^f\}$ from the spatial and frequency domains.
Inspired by previous works \cite{han2020trusted, han2022trusted}, we apply Dempster's combination rule \cite{sentz2002combination} to get the joint detection probability mass set $\widetilde{M}$ in the following manner: $\widetilde{M}=M^s\oplus M^f$.
The specific calculation rules for belief mass and uncertainty mass are formulated as
\begin{equation}
\widetilde{b}_k=\gamma(b_k^sb_k^f+b_k^su^f+b_k^fu^s),\,\,\, \widetilde{u}=\gamma u^su^f,
\end{equation}
where $\gamma=1/({1-\sum_{i\neq j}b_i^sb_j^f})$ is the scaling factor, normalizing the mass fusion to mitigate the effects of conflicting information between the spatial and frequency masses.
With this newly obtained joint detection mass $\widetilde{M}$, the joint evidence and the parameters of the Dirichlet distribution are calculated as follows:
\begin{equation}
    \widetilde{S}=\frac{K}{\widetilde{u}},\,\,\, \widetilde{e}_k=\widetilde{b}_k\times \widetilde{S},\,\,\,{\rm and}\,\,\,\, \widetilde{\alpha}_k=\widetilde{e}_k+1.
\end{equation}
For a test sample $x^i$, the model prediction $\widetilde{p}_k^i$ for class $k$ is computed as $\widetilde{p}_k^i=\widetilde{\alpha}_k^i/\widetilde{S}^i$. 

\begin{algorithm}[tbp]
\caption{DLED Inference Procedure}
\label{alg:algorithm}
\begin{algorithmic}[1]
\REQUIRE Input image $x$; uncertainty thresholds $\{\tau_k\}_{k=1}^K$ for each known class.
\STATE Obtain the frequency map: $x^f = \mathrm{FFT}(x)$
\STATE Extract dual-level evidence: $e^s = h(\mathcal{F}^s(x))$, $e^f = h(\mathcal{F}^f(x^f))$
\STATE Calculate belief and uncertainty for each branch: \\
\hspace{1em} $S = \sum_{k=1}^K (e_k + 1)$, \quad $b_k = e_k / S$, \quad $u = K / S$
\STATE Fuse belief and uncertainty: \\
\hspace{1em} $\widetilde{b}_k = \gamma (b_k^s b_k^f + b_k^s u^f + b_k^f u^s)$, \\
\hspace{1em} $\widetilde{u} = \gamma u^s u^f$, \quad $\gamma = 1 / \left( 1 - \sum_{i \neq j} b_i^s b_j^f \right)$ \hfill [Eq.~5]
\STATE Calculate fused evidence and parameters: \\
\hspace{1em} $\widetilde{S} = K / \widetilde{u}$, \quad $\widetilde{e}_k = \widetilde{b}_k \times \widetilde{S}$, \quad $\widetilde{\alpha}_k = \widetilde{e}_k + 1$ \hfill [Eq.~6]
\STATE Calculate predictive distribution and predicted label: \\
\hspace{1em} $\widetilde{p}_k = \widetilde{\alpha}_k / \widetilde{S}$, \quad $\widetilde{y} = \arg\max_k \widetilde{p}_k$
\STATE Improve uncertainty estimation: \\
\hspace{1em} $\hat{u} = 1 / \left( \max \{ \widetilde{\alpha}_1, \dots, \widetilde{\alpha}_K \} \right)$ \hfill [Eq.~8]
\STATE Novel category detection and final prediction: \\
\hspace{1em} $\hat{y} = 
\begin{cases}
\widetilde{y}, & \text{if } \hat{u} \leq \tau_{\widetilde{y}} \\
K+1, & \text{if } \hat{u} > \tau_{\widetilde{y}}  
\end{cases}$\hfill
\RETURN $\hat{y}$ and $\hat{u}$
\end{algorithmic}
\end{algorithm}

\noindent\textbf{Improved Uncertainty Estimation.}
Considering $\widetilde{u}=K/\widetilde{S}$ and $\widetilde{S}=\sum_{k=1}^{K}(\widetilde{e}_k+1)$, after dividing numerator and denominator by $K$, the uncertainty can be expressed as
\begin{equation}
    \widetilde{u} = \frac{1}{1 + \frac{1}{K}\sum_1^K\{\widetilde{e}_{1,\dots,K}\}},
\end{equation}
which indicates that the uncertainty is assessed using the average evidence across all $K$ classes.
Therefore, when the input data shows high evidence from irrelevant classes, the estimated uncertainty will be overestimated, resulting in a sub-optimal estimation.
To solve this problem, we propose an improved uncertainty estimation by replacing the \emph{average} evidence with \emph{maximum} evidence:
\begin{equation}
\label{Eq:improved_uncertainty}
    \hat{u}=\frac{1}{1+\max\{\widetilde{e}_{1,\dots,K}\}}=\frac{1}{\max\{\widetilde{\alpha}_{1,\dots,K}\}},
\end{equation}
where $\{\widetilde{e}_{1,\dots,K}\}$ represents the set of $K$ fused evidence. 
Our improved uncertainty measure offers the advantage of being less affected by low-evidence classes while retaining a normalized range between 0 and 1 for better human understanding. 
Moreover, it directly reflects the model’s confidence in the predicted class.
We recalculate the uncertainty $\hat{u}$ with (\ref{Eq:improved_uncertainty}) after the evidence fusion to achieve better detection performance.

To determine if a face image belongs to an unseen fake category, our model compares its uncertainty $\hat{u}$ with the uncertainty threshold for its predicted class. 
If the uncertainty exceeds the threshold, the model reassigns the label to the novel category.
We provide a detailed description of the inference procedure for our DLED model.
We denote the spatial and frequency branches as $\mathcal{F}^s$ and $\mathcal{F}^f$, respectively, such that $\mathcal{F} = \{\mathcal{F}^s, \mathcal{F}^f\}$.  
Given the retrieved uncertainty thresholds $\{\tau_k\}_{k=1}^K$ for each known class $k$, the DLED inference procedure for an input face image $x$ is described in Algorithm~\ref{alg:algorithm}, where $h(\cdot)$ denotes the evidence function and $\rm{FFT}(\cdot)$ represents the Fast Fourier Transform.
\section{Experiments}
\subsection{Experimental Setup}
\noindent{\bf Datasets.}
To evaluate model performance on the OSFFD problem, we conducted experiments using the comprehensive dataset DF40 \cite{df40}.
DF40 contains fake faces from four distinct categories (``Face Swapping", ``Face Reenactment", ``Entire Face Synthesis", and ``Face Editing") and includes a total of 40 diverse forgeries.
Additionally, we introduced data from two ``Stacked Manipulation" (SM) forgeries \cite{he2021forgerynet}, in which techniques from multiple fake categories are applied within a single image. 
We treat these SM forgeries as an auxiliary fake category.

\noindent{\bf Evaluation Protocols.} 
In the OSFFD problem, the training set comprises real faces and fake faces from multiple known fake categories, while the test set additionally includes samples from unknown fake categories. 
To evaluate the model’s performance, we first adopted the leave-one-out strategy in which one fake category from FS, FR, or EFS was withheld during training and treated as an unseen category during testing.
Subsequently, all three fake categories (FS, FR, and EFS) were included as seen classes, and the model was evaluated on a test set containing additional forgeries from FE and SM, representing novel fake categories.
As for the evaluation metrics, we employed the multiclass classification Accuracy ({\bf Acc}) and the Detection Rate ({\bf DR}), where DR refers to the recall of the unseen fake categories.

We compared the DLED method with the following baselines.
1) Two-stage baselines: 
We introduced a second training stage for one-class out-of-distribution (OOD) detection methods: OC-FakeDetect \cite{oc-dect} and SBI \cite{sbi}, in which an additional closed set multiclass model is independently trained to further classify the seen classes.
For fair comparison, we used CLIP as the multiclass model's backbone and finetuned it in the closed set manner with the cross-entropy loss;
2) CNN-based baselines: Xception \cite{ff++}, SPSL \cite{spsl}, SIA \cite{sia}, UCF \cite{ucf}, and NPR \cite{npr};
3) CLIP-based baselines: Zero-shot CLIP \cite{clip} and three established methods UnivFD~\cite{UniFD}, CLIPing \cite{prompt_tuning}, and $D^3$ \cite{D3}.

For the two-stage baselines, images recognized by the one-class model as seen classes are passed to the multiclass model to obtain their specific labels during testing.
For the CNN-based and CLIP-based baselines, we replaced their binary classifier with a multi-class classifier trained end-to-end, and adopted the MaxLogit \cite{zhang2023decoupling} technique during testing, due to its strong performance in detecting unknown samples.
For all algorithms that need a threshold to detect novel categories, we computed it from the training data such that 95$\%$ of the samples in each class are marked as known, which is the widely used setup in open set problems. 

\noindent{\bf Network Modules.} 
Our DLED method employs CLIP \cite{clip} as the backbone for both the spatial and frequency branches, using the ViT-B/16 \cite{vit} encoder with half-precision (fp16) data type.
The same CLIP configuration is also used for all CLIP-based baseline methods to ensure fair comparisons.
For the CLIP Zero-Shot baseline, we used the text prompt ``a [CLASS] photo." as the input, where “[CLASS]” is replaced with the predefined class names corresponding to the seen classes.
For CNN-based baseline methods and one-class detectors in Two-stage baselines, all backbones utilize the official implementations provided by the respective authors. 
These baseline models are trained from scratch for the OSFFD experiments.
Specifically, SPSL \cite{spsl} and UCF \cite{ucf} utilized the Xception backbone \cite{ff++}, while SBI \cite{sbi} and SIA \cite{sia} are based on EfficientNet-B4 \cite{efficientnet}. 
NPR \cite{npr} is a modified version of ResNet \cite{he2016deep}, and OC-FakeDetect \cite{oc-dect} incorporates a Variational Autoencoder \cite{kingma2013auto}. 
Xception \cite{ff++} is used directly without modification.

\noindent{\bf Implementation Details.}
In our experiments, the batch size was set to 32 for training and 100 for testing across all models.
Our DLED model employed the softplus function as its confidence function $h(\cdot)$.
Both DLED and CLIP-based baseline models are optimized using SGD with a learning rate of 0.002, while the remaining baseline methods use the Adam optimizer with a learning rate of $1e^{-4}$. 
All models, except for the CLIP Zero-Shot baseline, are trained for 50 epochs.
We randomly selected a seed and evaluated the model’s performance at the final epoch in a single run.
All methods are implemented in PyTorch, and experiments are conducted on an RTX 6000 Ada GPU.

\subsection{Evaluation of Detection Performance}
\noindent{\bfseries Open Set Face Forgery Detection.}
\begin{table}
  \centering
  \footnotesize
  \setlength{\tabcolsep}{3.2mm}        
  \caption{
    Comparisons of prediction accuracy with diverse baselines implemented by ourselves for the Real-vs-Fake detection task.
    Data configurations are the same as those in OSFFD.
    All baseline models are implemented following their original algorithms.
  }
  \label{tab:main_RvF}
  \begin{adjustbox}{width=\linewidth}
  \begin{tabular}{cccccc}
    \hline
    Methods                   & FS    & FR    & EFS   & FE \& SM & Avg   \\ \hline
    OC-FakeDect \cite{oc-dect} & 48.09 & 48.45 & 48.18 & 47.16   & 47.97 \\
    SBI \cite{sbi}                & 50.13 & 50.36 & 50.07 & 49.96   & 50.13 \\ \hline
    Xception \cite{ff++}        & 71.73 & 67.98 & 67.19 & 67.49   & 68.60 \\
    SPSL \cite{spsl}              & 72.29 & 65.87 & 70.34 & 69.57   & 69.52 \\
    SIA  \cite{sia}               & 69.45 & 64.13 & 66.91 & 64.64   & 66.28 \\
    UCF \cite{ucf}                & 71.10 & 64.78 & 65.18 & 67.98   & 67.26 \\
    NPR \cite{npr}                & 80.76 & 75.73 & 77.67 & 77.21   & 77.84 \\ \hline
    CLIP Zero-Shot \cite{clip}  & 52.96 & 53.20 & 53.12 & 56.62   & 53.97 \\
    UnivFD \cite{UniFD}           & 77.64 & 76.83 & 79.33 & 81.31   & 78.78 \\
    CLIPing \cite{prompt_tuning}  & 78.46 & 77.15 & 79.58 & 81.09   & 79.07 \\
    $D^3$ \cite{D3}               & 78.56 & 77.00 & 79.67 & 79.81   & 78.76 \\ \hline
    \textbf{Ours}                  & \textbf{87.22} & \textbf{85.93} & \textbf{83.52} & \textbf{84.97} & \textbf{85.41} \\ \hline
  \end{tabular}
  \end{adjustbox}     
  \end{table}
Since the two-stage baselines rely on the closed set finetuned CLIP model as their multiclass classifier, we also report the performance of this model independently.
As shown in Table \ref{tab:main_OSFFD}, most baseline models struggle to achieve high performance on both Accuracy (Acc) and Detection Rate (DR) simultaneously. 
Methods with higher Acc typically exhibit lower DR, and vice versa. 
It can also be observed that two-stage methods yield lower Acc than their base forgery classifier (CLIP Finetuning), indicating that OOD detectors and forgery classifiers are difficult to integrate in OSFFD with satisfactory performance. 
Additionally, directly applying OSR techniques with an Xception backbone attains notably low Acc, underscoring that off-the-shelf OSR approaches are insufficient to solve the OSFFD problem.
With more sophisticated designs tailored to face forgery detection, the baselines achieve higher Acc in most cases, confirming that efficient mechanisms for exploring forgery-specific representations are necessary to address OSFFD.
In comparison, our DLED model consistently achieves the highest DR across all scenarios and demonstrates superior average Acc, outperforming baseline methods in the majority of cases.
These results highlight the effectiveness of DLED in discovering novel fake categories while maintaining strong recognition performance on real images and known fake categories.

\noindent{\bfseries Real-vs-Fake Detection.}
We also evaluate the proposed DLED model on the traditional Real-vs-Fake detection task, using the same data configuration as in the OSFFD problem. 
In this task, all baseline methods are implemented according to their original designs without modification. 
For our DLED model, any face predicted to belong to a fake category is classified as a fake sample.
The results are shown in Table \ref{tab:main_RvF}.
It can be observed that DLED significantly outperforms these face forgery detection algorithms across all evaluation cases.
These empirical results demonstrate that, in addition to its strong performance on the OSFFD problem, the proposed DLED model also achieves competitive results on the traditional binary Real-vs-Fake deepfake detection task.

\subsection{Ablation Study}
In this section, we conducted an ablation study on DLED.
These experiments follow the same setup as described for OSFFD, and the results are summarized in Table \ref{tab:ablation}.
\begin{table}
  \centering
  \caption{
    Ablation Study of DLED.
    The table presents DR results under the same data configuration as used in the main OSFFD experiments.
  }
  \label{tab:ablation}
  \begin{adjustbox}{width=\linewidth}
  \begin{tabular}{ccccccc}
    \hline
    \multicolumn{2}{c}{Models} & FS & FR & EFS & FE \& SM & Avg \\ \hline
    \multicolumn{1}{c|}{\multirow{3}{*}{Spatial Branch}} &
      Zero-Shot with MaxLogit & 0.81 & 0.26 & 0.38 & 0.25 & 0.42 \\
    \multicolumn{1}{c|}{} &
      Zero-Shot with EDL      & 1.58 & 0.58 & 0.68 & 0.63 & 0.87 \\
    \multicolumn{1}{c|}{} &
      Finetuning with EDL     & 13.02 & 30.94 & 8.33 & 50.59 & 25.71 \\ \hline
    \multicolumn{1}{c|}{\multirow{3}{*}{Frequency Branch}} &
      Zero-Shot with MaxLogit & 3.85 & 2.35 & 6.98 & 0.53 & 3.43 \\
    \multicolumn{1}{c|}{} &
      Zero-Shot with EDL      & 4.71 & 2.51 & 6.06 & 0.55 & 3.46 \\
    \multicolumn{1}{c|}{} &
      Finetuning with EDL     & 14.34 & 8.49 & 7.69 & \textbf{90.36} & 30.22 \\ \hline
    \multicolumn{1}{c|}{\multirow{2}{*}{Two Branches}} &
      Evidence Fusion         & 32.42 & \textbf{36.16} & 32.56 & 79.74 & 45.22 \\
    \multicolumn{1}{c|}{} &
      Full DLED               & \textbf{33.61} & 34.92 & \textbf{34.71} & 82.18 & \textbf{46.36} \\ \hline
  \end{tabular}
  \end{adjustbox}
  \vspace{-6pt}
  \end{table}
Our results indicate that:
1)
Compared to MaxLogit, EDL enhances model performance across both the spatial and frequency branches, indicating its superior capability in uncertainty estimation and, consequently, improved discovery of novel categories.
2) 
Although equipped with EDL, the pretrained CLIP model cannot be directly applied to the OSFFD problem in either the spatial or frequency domain, as indicated by its extremely poor performance (see the 2nd and 5th rows).
Fine-tuning the prompts and integrating LoRA layers substantially improves the performance of both branches, highlighting the effectiveness of task-specific representation adaptation.
3) 
Without frequency information, the finetuned spatial branch with EDL exhibits an average performance drop of about 20$\%$ relative to the fused model (see the 3rd and 7th rows). 
This highlights the necessity of extracting complementary evidential cues across spatial and frequency domains to fully exploit forgery-specific signals and make more effective use of EDL, as well as the benefits of evidence integration.
4) 
By incorporating the improved uncertainty estimation, the full DLED model achieves the highest average Detection Rate, surpassing simple evidence fusion in most cases and thereby validating its effectiveness.

\subsection{Analysis of Visual Attention and Evidence}
\begin{figure}[b]
\centering
\includegraphics[width=\linewidth]{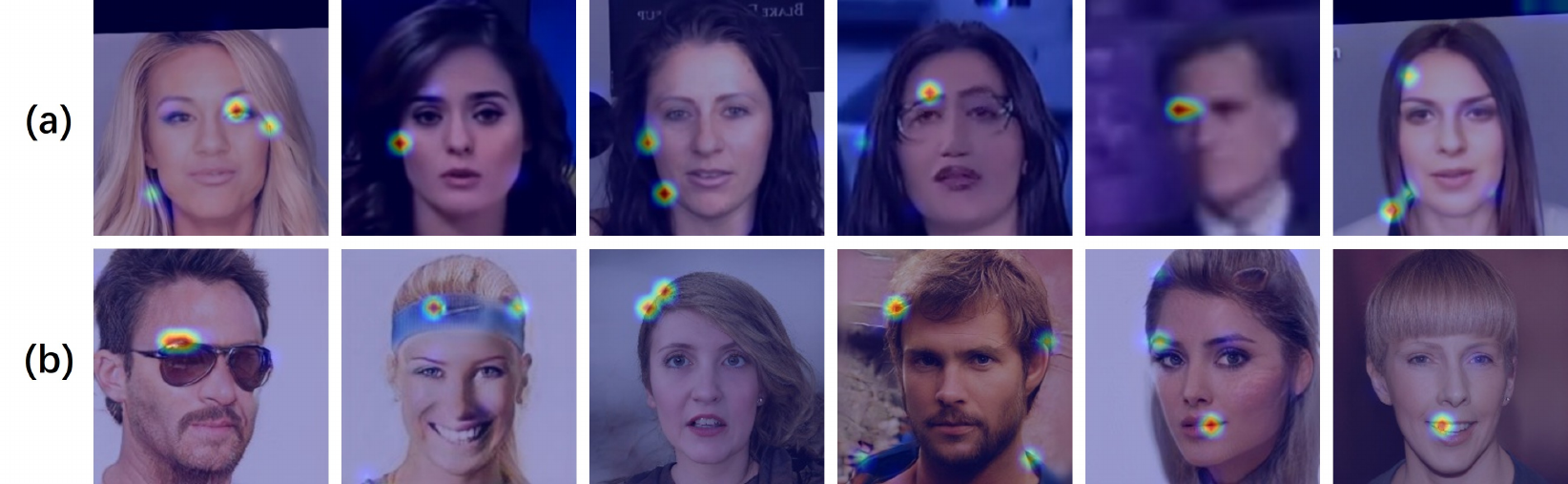}
  \caption{
  {\bfseries Visualization of Attention Map.}
  Attention maps generated by the DLED model for the novel fake categories FS and FE are shown in subfigures (a) and (b), respectively.
  }
  \label{fig:atten_map}
\end{figure}

\begin{figure}[tbp]
  \centering
  \includegraphics[width=\linewidth]{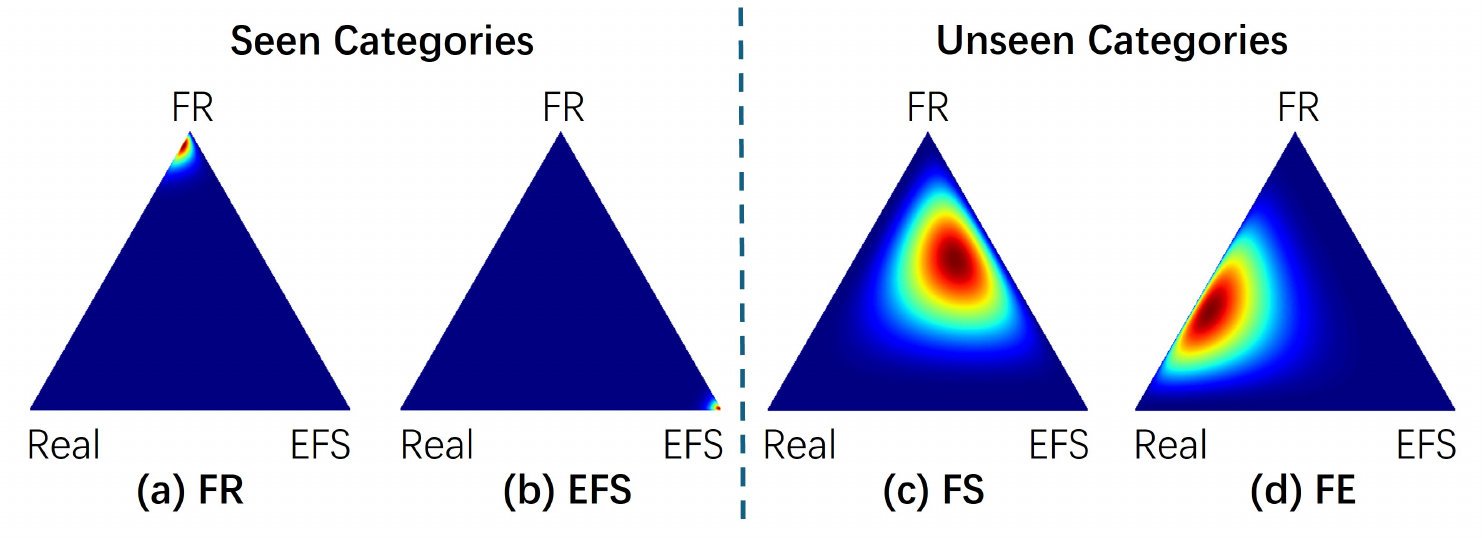}
  \caption{\textbf{Visualization of Evidence Distribution.}
  The evidence for seen fake categories FR and EFS is condensed in their corresponding corner with low uncertainty, while the evidence for novel fake categories FS and FE is sparse with higher uncertainty.}
  \label{fig:vis1}
  \end{figure}

\begin{figure}[htbp]
\centering
\includegraphics[width=\linewidth]{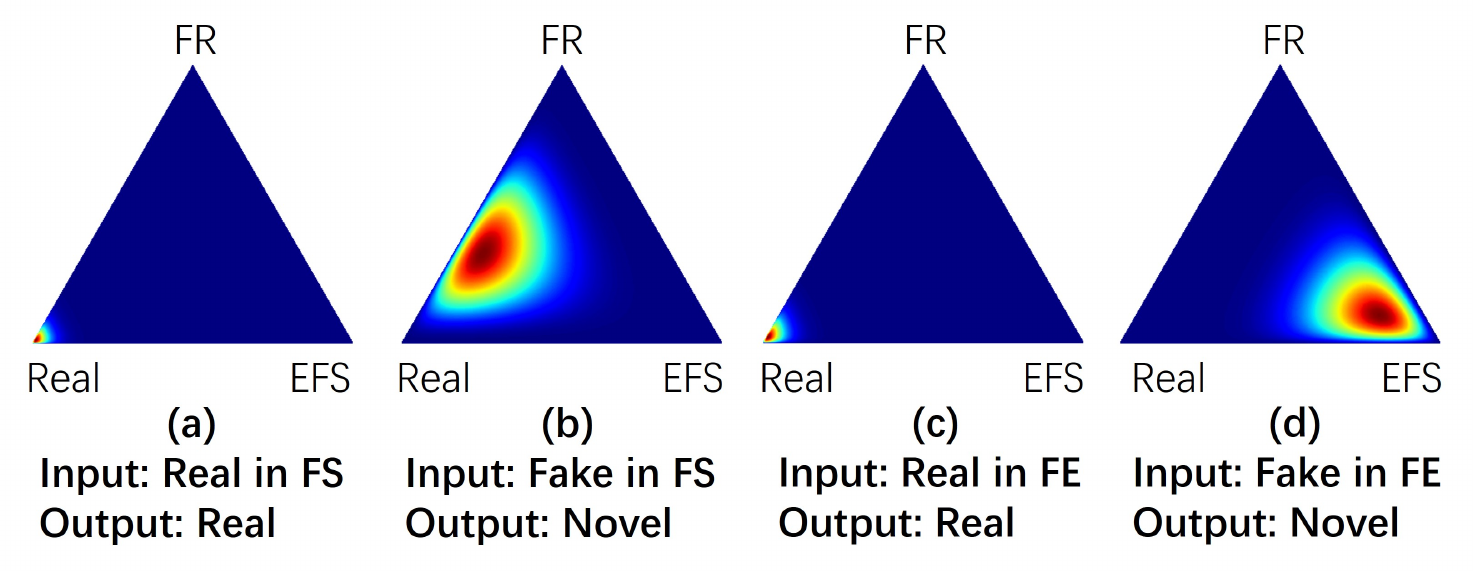}
  \caption{
  {\bfseries Visualization of Evidence Distribution for novel real and fake face images.}
 We visualize the Evidence Distribution of novel real and fake faces and present the corresponding predictions made by our DLED model.
 The prediction confidence for new real faces remains high, whereas novel fake faces exhibit low confidence accompanied by high prediction uncertainty.
  }
  \label{fig:vis2}
\end{figure}

To provide a clearer understanding of DLED's behavior in the OSFFD problem, we present visualizations of attention maps in Fig. \ref{fig:atten_map} and evidence distribution in Fig. \ref{fig:vis1} and Fig. \ref{fig:vis2}. 
In this analysis, FR and EFS are treated as known fake categories, while FS and FE represent novel categories. 

Fig.~\ref{fig:atten_map} illustrates the evidence collected by the DLED model. 
It can be observed that DLED attends to category-specific semantic cues. 
For example, in the case of FS, the model highlights edge regions indicative of face transplantation, whereas for FE, it focuses on manipulated areas such as sunglasses, hairbands, and hair.

Fig.~\ref{fig:vis1} illustrates how uncertainty estimation facilitates the detection of novel fake categories among test samples. 
Each subfigure visualizes the Dirichlet distribution produced by DLED for the corresponding fake category.
These visualizations demonstrate that the DLED model exhibits higher confidence when making predictions on seen classes, while showing greater prediction uncertainty for novel fake categories.  
This behavior enables DLED to effectively recognize newly emerging fake categories while simultaneously maintaining strong performance on known classes.
Furthermore, we illustrate how DLED distinguishes between unseen real and unseen fake faces. 
We sample unseen real and fake images from the novel categories FS and FE, respectively, and visualize their individual evidential distributions in Fig.~\ref{fig:vis2}.
These visualizations show that the DLED model can produce low uncertainty scores for real images, even though they are from the unseen FS and FE categories, and produce high uncertainties for fake images since they are novel fakes.

\subsection{Influence of Uncertainty Threshold}
\begin{figure}[htbp]
    \centering
\includegraphics[width=0.89\linewidth]{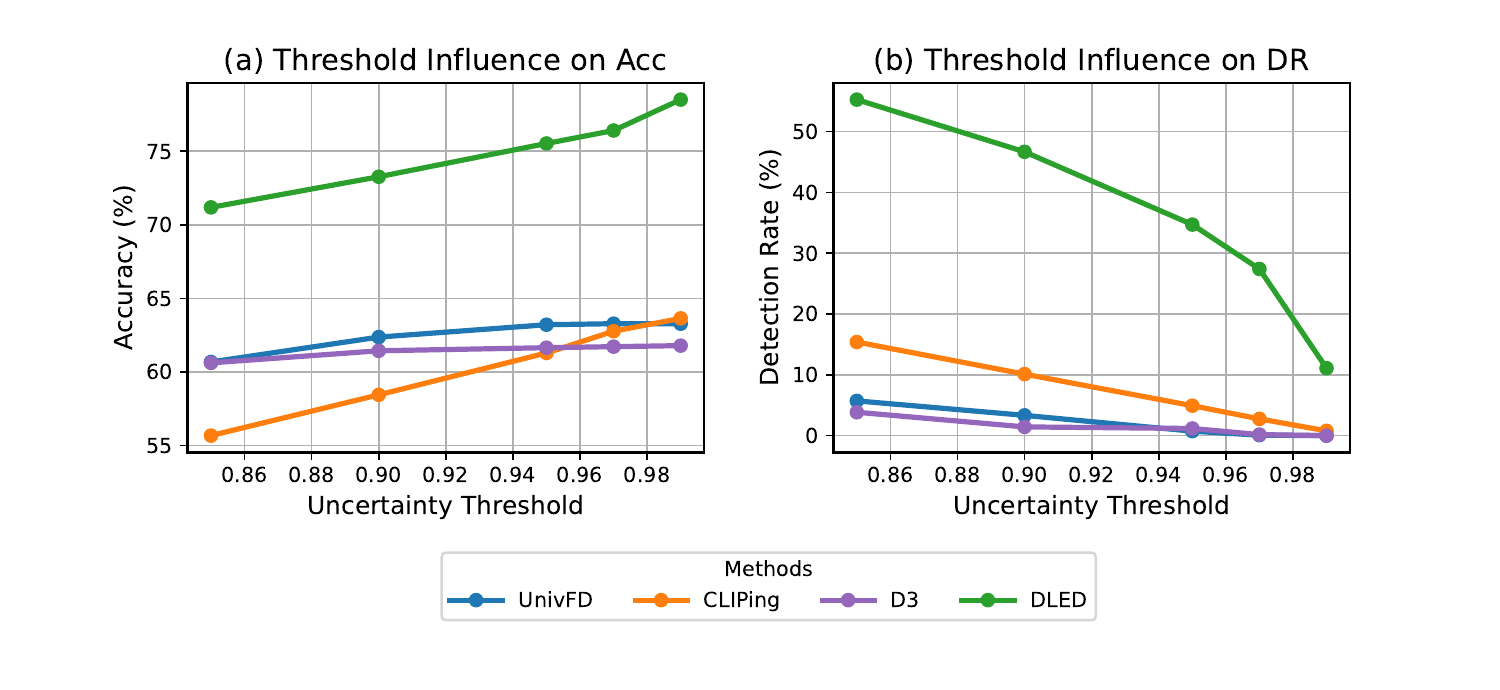}
    \caption{Accuracy and Detection Rate across different uncertainty thresholds for UnivFD, CLIPing, and DLED.}
    \label{fig:combined_results}
\end{figure}
The uncertainty threshold is calculated from the training data such that a certain percentage of the samples in each class is marked as known.
A higher threshold indicates that more training samples are regarded as reliable, thereby imposing a stricter criterion for novel fake discovery.
To further investigate the impact of the uncertainty threshold on the OSFFD problem, we conducted a series of experiments using various threshold values: [0.85, 0.90, 0.95, 0.97, 0.99] on DLED, as well as three CLIP-based baseline methods: UnivFD \cite{UniFD}, CLIPing \cite{prompt_tuning}, and $D^{3}$ \cite{D3}.

In these experiments, FS and FR are treated as seen fake categories, while EFS is designated as the only novel fake category.
The results are presented in Fig.~\ref{fig:combined_results}.
It can be observed that as the uncertainty threshold increases, classification accuracy improves, whereas the novel fake discovery rate declines. 
This indicates that a stricter threshold reduces the detection of novel samples but also decreases the misclassification of known samples. 
These findings highlight the critical role of the uncertainty threshold in balancing known class classification and novel fake discovery.
Moreover, across all threshold settings, our DLED method consistently achieves higher Acc and DR compared to the three CLIP-based baselines, demonstrating its robustness and superior performance.
\section{Conclusion}
In this work, we reformulate the Open Set Face Forgery Detection (OSFFD) problem by removing the need for unlabeled novel data during model training, thereby enhancing its practicality for real-world applications.
By treating OSFFD as an uncertainty estimation problem, we propose a novel algorithm, DLED, which effectively identifies unseen fake categories as novel while simultaneously classifying real and known fake categories. 
DLED leverages EDL to collect and fuse evidence from both the spatial and frequency domains, exploiting category-specific semantics to estimate prediction uncertainty. 
Additionally, we propose an improved uncertainty formulation that enhances the model's ability to detect novel fake categories.
Extensive experiments under various testing configurations demonstrate that DLED substantially outperforms diverse baseline methods in addressing the OSFFD problem.
Future work will focus on improving the efficiency of the proposed method and enabling rapid adaptation to the detected novel fake categories.

\clearpage
{\small
\bibliographystyle{ieee}
\bibliography{egbib}
}

\end{document}